\documentclass[10pt,journal,cspaper,compsoc]{IEEEtran}
\usepackage[utf8]{inputenc}
\ifCLASSOPTIONcompsoc
\usepackage[nocompress]{cite}
\else
\usepackage{cite}
\fi
\usepackage{caption}
\usepackage{graphicx}
\usepackage{subcaption}
\usepackage{epsfig}
\usepackage{amsmath}
\usepackage{amssymb}
\usepackage{cases}
\usepackage{hhline}
\usepackage{indentfirst} 
\usepackage{algorithm}
\usepackage{algorithmic}
\usepackage[T1]{fontenc}
\usepackage{color}

\begin{document}
	\title{Federated Learning for Data Market: Shapley-UCB for Seller Selection and Incentives}
	\author{Kongyang Chen, Zeming Xu
		\IEEEcompsocitemizethanks{
			\IEEEcompsocthanksitem K. Chen is with the School of Intelligent Systems Engineering, Sun Yat-Sen University, Guangzhou 510275, China. K. Chen is also with the Department of Computing, The Hong Kong Polytechnic University, Hong Kong, China. E-mail: kongyang.chen@gmail.com.
	}}

	\IEEEtitleabstractindextext{
		\begin{abstract}
			In recent years, research on the data trading market has been continuously deepened. In the transaction process, there is an information asymmetry process between agents and sellers. For sellers, direct data delivery faces the risk of privacy leakage. At the same time, sellers are not willing to provide data. A reasonable compensation method is needed to encourage sellers to provide data resources. For agents, the quality of data provided by sellers needs to be examined and evaluated. Otherwise, agents may consume too much cost and resources by recruiting sellers with poor data quality. Therefore, it is necessary to build a complete delivery process for the interaction between sellers and agents in the trading market so that the needs of sellers and agents can be met. The federated learning architecture is widely used in the data market due to its good privacy protection. Therefore, in this work, in response to the above challenges, we propose a transaction framework based on the federated learning architecture, and design a seller selection algorithm and incentive compensation mechanism. Specifically, we use gradient similarity and Shapley algorithm to fairly and accurately evaluate the contribution of sellers, and use the modified UCB algorithm to select sellers. After the training, fair compensation is made according to the seller's participation in the training. In view of the above work, we designed reasonable experiments for demonstration and obtained results, proving the rationality and effectiveness of the framework.
		\end{abstract}
		
	}	
	\maketitle
	\IEEEdisplaynontitleabstractindextext
	\IEEEpeerreviewmaketitle
	\section{Introduction}\label{sec:introduction}
	\subsection{Background}
In the classic federated learning architecture \cite{1}, the local client provides data for model training. After each round of global training, the client sends the model to the central server for model aggregation. The central server does not need To collect data, only a subset of clients are randomly sampled in each global training, so federated learning has good data privacy. In this assumption, the central server does not need to care about whether there are clients willing to provide data for training, nor does it need to participate in client selection and consider the client's training costs and returns. However, this is not practical in real situations, especially In the transaction market, the willingness of sellers to provide data largely depends on the degree of compensation that agents provide to sellers.
In recent years, federated learning has received widespread attention due to its performance and better privacy. In federated learning scenarios, the problem of how agents recruit high-quality sellers to participate in task training and provide reasonable compensation has not been effectively solved. Under normal circumstances, it is difficult to collect data directly. The federated learning architecture can avoid directly obtaining data for model training, and use the model parameters obtained from local data training to upload and update to obtain the final model. In this process, recruiting sellers is an information asymmetric process. First, the seller's data quality is invisible to the agent. Feedback can only be provided through the accuracy of the model and the corresponding evaluation algorithm during the training process. If a batch of data is recruited, Sellers with poor quality will lead to poor quality of the final model, which is not conducive to agents delivering models, and will also cause excessive consumption of training time and training resources. Second, even if there are high-quality sellers willing to provide data, how to select sellers is a complicated process. Due to the lag in the contribution evaluation method, it is difficult for agents to accurately identify high-quality sellers at the beginning of training. Third, the rewards brought by the data provided by the seller are not reasonable. Essentially, after the training, the agent cannot make a reasonable evaluation based on the seller's contribution and provide reasonable compensation based on the contribution, resulting in the inability to motivate the seller to provide data resources. , which is not conducive to the development of the data trading market environment.
For one thing, it is important to design a reasonable contribution evaluation method. Many studies have provided methods to evaluate the value of data. For example, Song et al. \cite{2} used the gradient generated by the client's local training of the temporary global model to reconstruct the new model, so as to calculate the shapley value of the client, effectively reducing the computing overhead. Jiang et al. \cite{3} proposed to estimate the contribution of clients in gradient and data space. Specifically, they suggested the difference in gradient direction between clients in gradient space, and the use of auxiliary models to measure the prediction error of client data in data space.
Regarding the second one, seller selection is an information asymmetric process. The reason is that there is a lag in the seller's contribution assessment during the training process. As a result, the agent cannot estimate the contribution probability of each seller in advance, and therefore cannot reasonably select a group of sellers to participate. train. There are many studies on client selection in federated learning. The selection metrics and application scenarios discussed in most studies are different. Ma et al. \cite{4} introduced a group-based client selection mechanism, based on the group bulldozer distance (GEMD) As a metric that divides clients into different groups to balance the client's local label distribution, a smaller GEMD value means that the client's local data is closer to the IID distribution. Ami et al. \cite{5} modeled the client selection problem as a multi-armed bandit (MAB) problem based on the client's training delay as an indicator to minimize the training delay.
For the third one, the focus is to design a reasonable incentive return mechanism. Compensation based on the degree of contribution is a reasonable way. Under ideal circumstances, sellers provide data voluntarily and do not require any compensation. However, in reality, resource consumption and The issue of privacy leakage exacerbates sellers' reluctance to provide data, and for agents, the failure to recruit high-quality clients means that the final delivery performance of the global model will be greatly affected. Therefore, designing a reasonable incentive mechanism is essential. There are many studies in this area, which have studied the design of incentive mechanisms under different indicators. Hu et al. \cite{6} designed an incentive mechanism based on game theory to select the best Users who can provide reliable data will be compensated reasonably based on the cost of privacy leakage. Sim et al. \cite{7} designed an incentive scheme based on the Shapley value and provided each customer with a customized model as compensation instead of a monetary reward. However, this scheme did not take into account the training cost of the model and cannot be directly applied to federation learning.
\subsection{Contribution}
In this paper, we focus on solving the above existing problems. We use existing research to design a reasonable contribution evaluation method, model seller selection as a MAB problem, redesign the reward function, and design a reasonable compensation function. Achieve reasonable compensation based on contribution. The main contributions of this work are as follows:
	\begin{enumerate}
		\item We propose to evaluate the training value of the model based on gradient similarity and Shapley value, combine the two to form the contribution evaluation value of each round, and design it as an aggregate parameter weight.	
		\item According to the unpredictability of the seller's performance during the training process, the seller selection problem is modeled as a MAB problem, and the ucb algorithm is used to use the contribution value as an indicator to redesign the reward function and update the selection probability based on the seller's historical performance.	
		\item Design a reasonable contribution aggregation function to aggregate the seller's training process contribution and local data distribution, and derive a reasonable compensation value based on budget constraints, ensuring that agents pay low costs and sellers receive high returns under this strategy.
	\end{enumerate}
	The rest of this paper is organized as follows. Chapter 2 will introduce the research content related to the content of this work. Chapter 3 will introduce the system architecture and main steps of this work. Chapter 4 will provide a detailed design of the technology involved in each step. Introduction, Chapter 5 will analyze the simulation experiment results, and Chapter 6 will summarize the content of this work.	
	\section{Preliminary Knowledge}\label{sec:Preliminary Knowledge}
	This chapter will introduce the work related to the research content of this work. Since this work is based on the federated learning architecture, it attempts to solve the demand problem between sellers and agents in the data trading market. Therefore, research on client contribution, client selection and incentive mechanisms in federated learning is relevant to this work, so this chapter will give a general introduction to the research in these aspects.
	\subsection{Client Contribution Assessment}
	Client contribution evaluation research aims to provide client selection metrics for client selection and incentive mechanism design. In machine learning, the measurement method based on the Shapley value \cite{8} is often used to evaluate the performance of the data pair model, based on the Shapley value The concept of Intermediate results are recorded and no additional model training is involved. Zhao et al. \cite{9} use the similarity between the local local model and the global aggregate model to measure the degree of contribution, assuming that the local model with greater contribution is more similar to the global model. Chen et al. \cite{10} measure the client's contribution by calculating the cross-entropy between each local model and the global aggregate model. In addition, Lyu et al. \cite{11} designed a reputation-based contribution evaluation method for fairness issues. The agent will iteratively update the reputation of each client based on the reputation value calculated in each round and its historical reputation.
	\subsection{Client Selection}
	Research on client selection focuses on designing reasonable and efficient selection algorithms to select high-quality clients, usually accompanied by evaluation of clients for screening. Ma et al. \cite{4} introduced a group-based client selection mechanism, dividing clients into different groups based on the group bulldozer distance (GEMD) as a metric to balance the local label distribution of the client. A smaller GEMD value means The client's local data is closer to the IID distribution. Ami et al. [5] modeled the client selection problem as a multi-armed bandit (MAB) problem based on the client's training delay as an indicator to minimize the training delay. Zhang et al. \cite{12} proposed a selection method based on reputation and reverse auction theory. After the agent releases the task, the client with good data quality is selected through the client's bidding and reputation. Hu et al. \cite{13} used game theory to model the utility maximization problem of servers and clients in federated learning as a two-stage Stackelberg game. This scheme derives the optimal strategies of servers and users by solving Stackelberg equilibrium, thereby selecting the most beneficial Clients that may provide reliable private data. Le et al. \cite{14} designed a selection method based on a random auction framework. Specifically, the client uploads bids based on local resources, and the agent selects the winning bidder based on the bidding information and algorithm.
	\subsection{Incentive Mechanism}
	The compensation design of the incentive mechanism usually chooses to compensate clients participating in training based on the revenue distribution method. Different revenue distribution methods are conducive to motivating participants with high-quality data to continue to participate in federated learning. \cite{15} proposed a hierarchical incentive mechanism that uses contract theory to establish an incentive mechanism between clients and agents. Zeng et al. \cite{16} proposed an incentive mechanism based on a multi-dimensional procurement auction framework, using game theory to derive the client's optimal strategy, thereby motivating more low-cost, high-quality clients to participate. Rehman et al. \cite{17} proposed a reputation system based on blockchain, which aggregates, calculates and records the reputation of each participant in federated learning through smart contracts, and motivates participants to produce honest and high-quality behaviors under reputation awareness. , thereby promoting the healthy development of the federated learning process. In addition, Sarikaya et al. \cite{18} analyzed the impact of heterogeneous data on the convergence speed of federated learning and proposed an incentive mechanism to balance the time delay of each iteration. Khan et al. \cite{19} used the theory of game theory to study enabling federated learning in edge networks, and established a specific incentive mechanism through Stackelberg game to motivate clients to participate in training.
	\section{System Model}\label{sec:System Model}
This chapter will introduce the system architecture of this work and the work content and technical applications involved in each step. The data transaction framework flow chart introduced in this article is shown in Figure \ref{fig:System Model}. It is mainly divided into the following nine steps:
	\begin{figure*}[!t]
		\centering
		\includegraphics[width=1.0\textwidth]{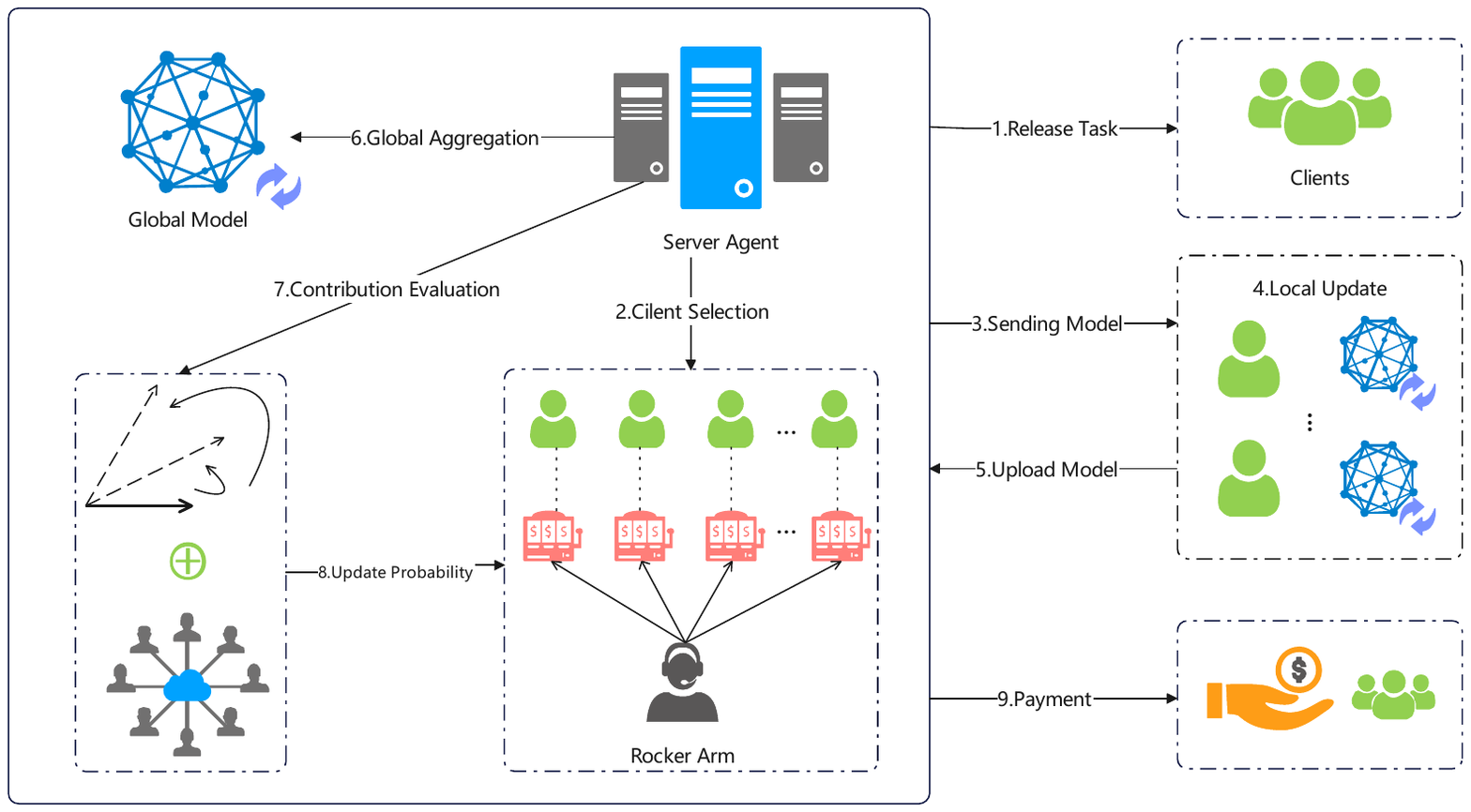}
				\vspace{-60pt}
		\caption{System Model}
		\label{fig:System Model}
	\end{figure*}
	\begin{enumerate}
		\item The agent publishes training tasks and recruits clients that meet the task data type.
		\item Selecting a client is divided into two parts:
Part One: In the first few rounds of training, clients are selected for training and all clients are evaluated once.
Part 2: After the first part, the agent will select the top m clients with the highest probability based on the selection probabilities formed by the clients’ past performance.
		
		\item Based on the selected set of clients, the agent sends the trained model to these clients.
		\item The client performs local model training.
		
		\item After each round of partial updates is completed, the client sends update parameters back to the agent.
		
		\item The agent performs aggregation and update of global model parameters.
		\item During the global model aggregation process, the agent will evaluate the client's performance in this round of training based on two aspects, including:
				Gradient direction information: Based on the cosine value of the local model gradient and the global model gradient, evaluate whether the local model can bring new direction information to the global model.
				Approximate Shapley value calculation: Save the gradient information of the local model during the training process, load different gradient information through the Shapley algorithm for accuracy evaluation, and calculate the Shapley value that reflects the amount of local data. A negative shapley value means that the local data quality is poor, and a positive shapley value means that the local data resources are better.
		\item After each round of global training, the agent uses the UCB algorithm to decide whether to reward the client based on the client's performance in this round, and updates the client's selection probability for the next round based on the reward probability and participation round.
		\item After the final model training is completed, the agent will provide reasonable compensation based on the training contribution of the recruited clients.
	\end{enumerate}
	\section{Detailed Design}\label{sec:Detailed Design}
	   This chapter introduces the detailed implementation of specific steps in the system architecture. Suppose the number of clients in reality is \textit{N},$ D^N=\{D_i\}^N_{i=1} $ represents the amount of local data owned by the client, and $ M^N=\{M_i\}^N_{i=1} $ represents the local training model of each client. The agent's task is to recruit suitable \textit{m} clients for each round of global training. The total number of training rounds is \textit{T}. After the training, the agent will make reasonable compensation based on the budget limit.
	\subsection{Contribution Evaluation Design} 
 This section introduces two factors for evaluating the client's contribution in a single round of training, considering the model gradient direction and the model's Shapley value respectively. We record the client's contribution evaluation value at the end of each training by multiplying the two and use this value as the aggregation parameter.
	\subsubsection{Gradient Direction Information}
 Gradient similarity is used to calculate the degree of proximity between two gradient directions. For the gradient update of the federated learning training process, different directions represent the degree of information that the local model can provide. The following calculation method is used to represent the impact of the local model on the global model.
\begin{equation}
\Gamma_{t,i} (\cos) \triangleq 1 - \cos(\nabla F_i(w_{t,i}), \nabla F_i(w_t^{-i}))
\end{equation}
 Where $\nabla F_i(w_{t,i})$ represents the gradient of the local model of client \textit{i} in round \textit{t}, and $\nabla F_i(w_t^{-i})$ represents the aggregated gradient of the global model composed of the remaining clients except client i in round \textit{t}. The cosine value between the two is calculated. If the value is closer to 1, the value of $\Gamma_{t,i} (\cos)$ is closer to 0, which means that the selection of local model \textit{i} cannot provide new directional information for the global model. On the contrary, when the cosine value is closer to 0, it means that the local model \textit{i} can provide more directional information. At this time, the value of $\Gamma_{t,i} (\cos)$ is closer to 1, and it is considered that the client can provide more update information. The value of $\Gamma_{t,i} (\cos)$ represents the importance of the gradient update of the model in round \textit{t}. The closer it is to 1, the more directional information the model can provide.
\subsubsection{Gradient Loading Approximation Shapley Algorithm}
 In federated learning, directly applying the Shapley algorithm will cause a huge computational burden. Since Shapley requires calculating all full permutations and evaluating them, and in federated learning, different full permutations correspond to complete local models or aggregate models composed of local models. Therefore, each retraining of the model will be seriously time-consuming and labor-intensive. In response to the above problems, some related articles have proposed to transform the original Shapley value to reduce the amount of calculation \cite{2,20,21}. This paper adopts a gradient loading-based method instead of model reconstruction training to reduce the amount of calculation. The specific Shapley value of the client is calculated through multiple rounds of aggregation. First, in each round of global training, the server will retain the gradient update information uploaded by each local model in this round. According to different client combinations, the global model loads the corresponding gradient information to form the corresponding aggregate model, and calculates and saves the performance of the aggregate model. Next, according to different aggregate models, the cumulative marginal contribution of each local model in this round is calculated. Finally, the cumulative marginal contribution is averaged to obtain the Shapley value of client \textit{i} in the \textit{t}th round, as follows, where \textit{S} represents a subset of $D^m$, and $D^m$ is the full set of \textit{m} clients.
\begin{equation}
 \phi_t^i = C \sum_{S \subseteq N \setminus \{i\}} \frac{U(M_{S \cup \{i\}}) - U(M_S)}{(n-1)\mathrm{!}/|S|}
\end{equation}
\subsubsection{Aggregation of Contributing Factors}
 In the classic federated learning algorithm FedAvg\cite{1}, when aggregating global models, the proportion of the local model's data set to the total data is used as the weight of the aggregation parameter. In our work, based on the two contribution evaluation factors introduced above, we redesign the weight to the normalized value of $\phi_t^i$. The aggregation weight of a local model i in round t is defined as:
\begin{equation}
p_t^i = \frac{\theta_t^i}{\sum_{i=1}^m \theta_t^i}
\end{equation}
 Then the global model aggregation parameters of round t are:
\begin{equation}
 w_{t+1} \leftarrow w_t - \eta \sum_{i=1}^m p_t^i \nabla F_i(w_{t,i})
\end{equation}
\subsection{Client Selection Design}
 Client selection is a problem worth studying. Different client selection algorithms are implemented based on different given scenarios. In this work, we start from the perspective of measuring client performance. At the beginning of a single round of training, the agent cannot accurately obtain the performance of each client in this round, which leads to the lag and unpredictability of the client's contribution evaluation. We model this problem as a MAB problem, that is, we use MAB-related algorithms to record the client's historical performance, so that at the beginning of each training, we can use the previous performance to evaluate the client's contribution in this round, and thus select the client that meets the requirements. In this work, we choose to use the UCB algorithm as the client selection algorithm, and redesign the reward function based on two contribution evaluation factors, as follows.
\subsubsection{UCB Algorithm}	
 The idea behind the upper confidence bound (UCB) algorithm selection is that the square root term is a measure of the uncertainty in the estimate of the contribution of the \textit{i}th model. Therefore, the size of the maximum value is an upper limit on the possible true value of model \textit{i}. Each time \textit{i} is selected, the uncertainty may decrease, and since $n_i$ appears in the denominator of the uncertainty term, this term decreases as $n_i$ increases. On the other hand, each time a seller other than \textit{i} is selected, \textit{n} in the numerator increases, while $n_i$ does not change, so the uncertainty increases. The use of the natural logarithm means that the increase becomes smaller and smaller over time, but it is infinite and all models will eventually be selected. Therefore, every seller has the opportunity to participate with a fair contribution, while ensuring that greedy seller selection will not occur, thereby avoiding the situation where the model generalization performance is poor.
 \begin{equation}
     \text{prob} = R_i + \sqrt{\frac{2 \ln n}{n_i}}
 \end{equation}
\subsubsection{Reward Function}
In the ucb formula, $R_i$ represents the probability of reward, which is reflected in the probability of generating a reward each time a selection is made. In this work, the reward is reflected in whether a client performs well or makes a positive contribution in this round of training when a client is selected for this round of training. Based on the two methods of measuring performance mentioned above, we design a corresponding reward function, which is defined as: the closer $\Gamma_{t,i} (\cos)$ of local model \textit{i} in the \textit{t}th round is to 1 (less than the threshold \textit{k}) and $\phi_t^i$ is greater than the average $\phi_t^i$ of the participating clients in the \textit{t}th round and the value is positive. When the above conditions are met, the reward value is 1, otherwise it is considered that the performance in this round is not worth rewarding, and the reward value is 0. $n_i$ represents the cumulative number of rounds participated by client \textit{i}, and \textit{n} represents all training rounds. Among them, \textit{k} is a pre-set threshold, and \textit{m} is the number of clients participating in each round.
\begin{equation}
R_i = \begin{cases} 
1, & \text{if } \Gamma_{(t,i)} (\cos) < k \cup \phi_t^i > 0 \cup \phi_t^i > \frac{\sum_{i=1}^m \phi_t^i}{m} \\
0, & \text{if } \neg \Gamma_{(t,i)} (\cos) < k \parallel \neg \phi_t^i > 0 \parallel \neg \phi_t^i > \frac{\sum_{i=1}^m \phi_t^i}{m}
\end{cases}
\end{equation}
	\subsection{Compensation Design}
	\subsubsection{Contribution Aggregate Function}
 When the global training round ends, in addition to obtaining the cumulative contribution evaluation value during the training process, it is still necessary to combine the client's local data distribution and the number of training times for investigation. Therefore, it is necessary to design a reasonable contribution aggregation function to obtain the final contribution value. This work will examine and measure from different aspects, including the participation round \textit{P}, the cumulative $\theta_t^i$ value represented by \textit{GS}, and the client's local data distribution \textit{EMD} and data volume \textit{C}, among which \textit{q} is the control function growth rate factor. The \textit{EMD} value is used to measure the similarity of two data distributions. In this work, this value is used to judge the degree to which the client's local data distribution deviates from the global data distribution. Therefore, a larger \textit{EMD} value represents a larger deviation of the client's local data distribution, so the difference of $1-EMD$ is taken as the value of this parameter. Among these four parameters, \textit{GS} is the most important measurement standard. If \textit{P} is used as the main indicator for measurement, the client's participation round will not be proportional to the actual contribution under the greedy selection participation round strategy, resulting in unfair contribution evaluation. Therefore, the weight of \textit{GS} should be larger, and the influence of \textit{P} should be suppressed, so that when the \textit{P} value is large, the impact on the function value is also within a controllable range.
The final contribution evaluation value of client \textit{i} is:
\begin{equation}
    P_i = \begin{cases} 
P_i + 1, & \text{if } i \subseteq m \text{ and } t > 0 \\
0, & \text{if } t = 0
\end{cases}
\end{equation}
\begin{equation}
    GS_{i} = \sum_{t=1}^T \theta_t^i
\end{equation}
\begin{equation}
    C_i = \frac{\textit{len}(D_i)}{\sum_{i=1}^N \textit{len}(D_i)}
\end{equation}
\begin{equation}
    \text{CE}_i = \left(1 + \log \left(P_i \cdot e^{GS_{i}} \right) \cdot \left[C_i + (1 - \text{EMD}_i)\right]\right) / e^{-q}
\end{equation}
	\subsubsection{Budget Constraint Compensation}
	After obtaining the client's contribution evaluation value, it is necessary to compensate the client fairly under the budget limit to ensure high income for the client. The contribution evaluation function ensures the fairness of the final contribution value. The final compensation value is calculated as follows:
 \begin{equation}
     CV_i = \frac{\textit{Budget} \cdot CE_i}{\sum_{i=1}^N CE_i}
 \end{equation}
\section{Performance Evaluation}\label{sec:Performance Evaluation}
	This chapter will design corresponding experiments to verify the research involved in this work. Five parts are designed for experiments and corresponding simulation results are provided to evaluate the effectiveness and rationality of the framework.
\subsection{Selection Strategy}
\begin{figure*}[!t]
    \centering
    \begin{subfigure}[!t]{\textwidth}
        \centering
        \includegraphics[width=0.8\textwidth]{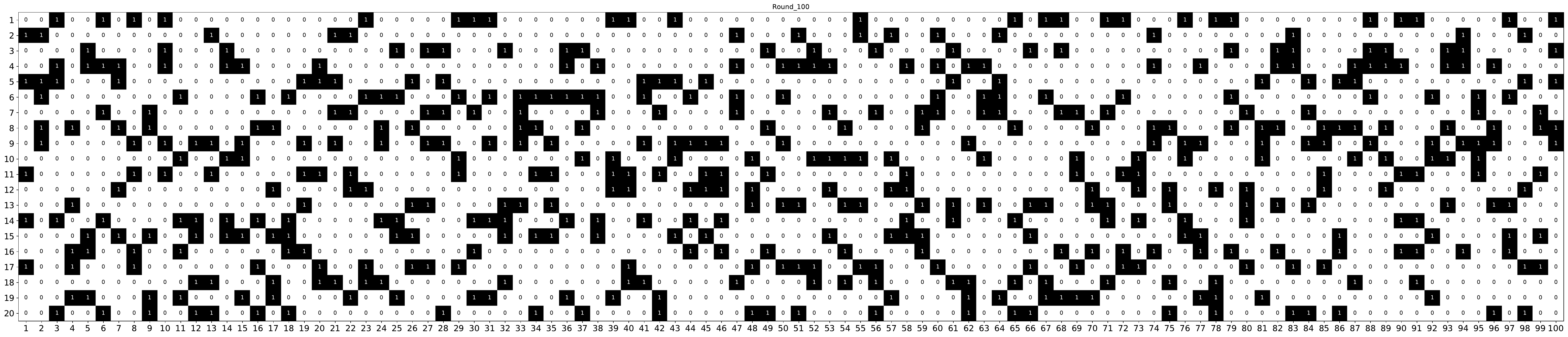}
        \caption{Random-Non-IID}
        \label{fig:random-NonIID}
    \end{subfigure}
    \begin{subfigure}[!t]{\textwidth}
        \centering
        \includegraphics[width=0.8\textwidth]{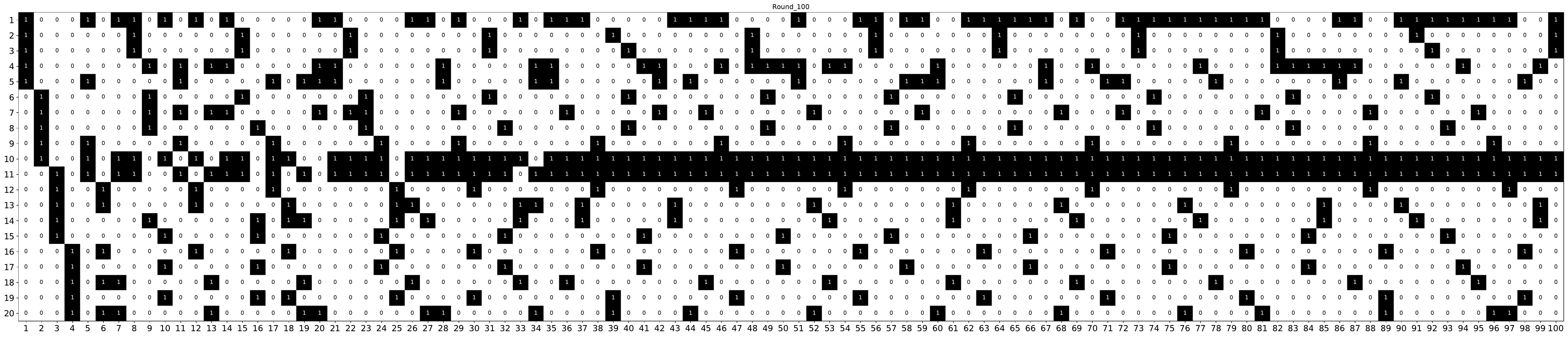}
        \caption{UCB-Non-IID}
        \label{fig:ucb-NonIID}
    \end{subfigure}
    \caption{Client Selection in Non-IID Case}
    \label{fig:non-iid}
\end{figure*}

\begin{figure*}[!t]
    \centering
    \begin{subfigure}[!t]{\textwidth}
        \centering
        \includegraphics[width=0.8\textwidth]{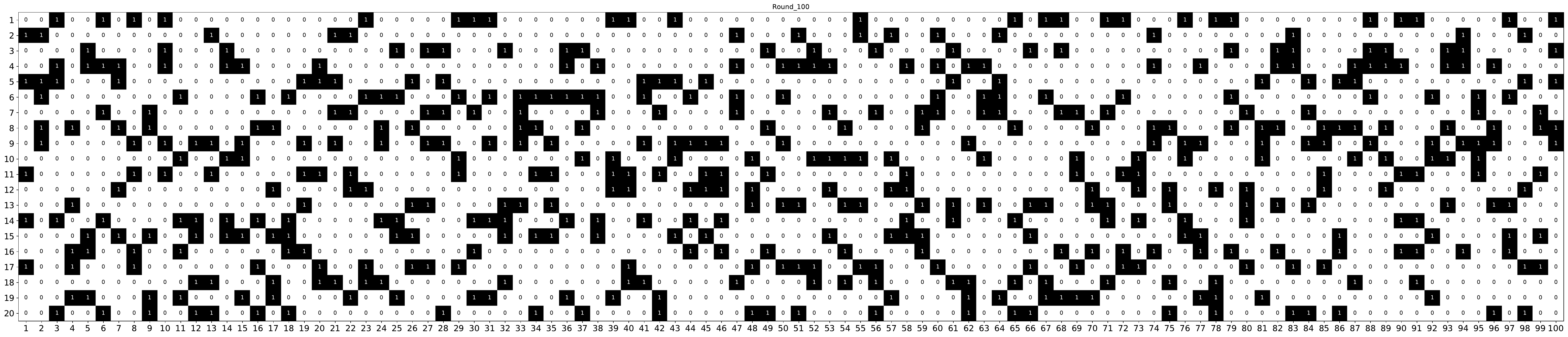}
        \caption{Random-IID}
        \label{fig:random-IID}
    \end{subfigure}
    \begin{subfigure}[!t]{\textwidth}
        \centering
        \includegraphics[width=0.8\textwidth]{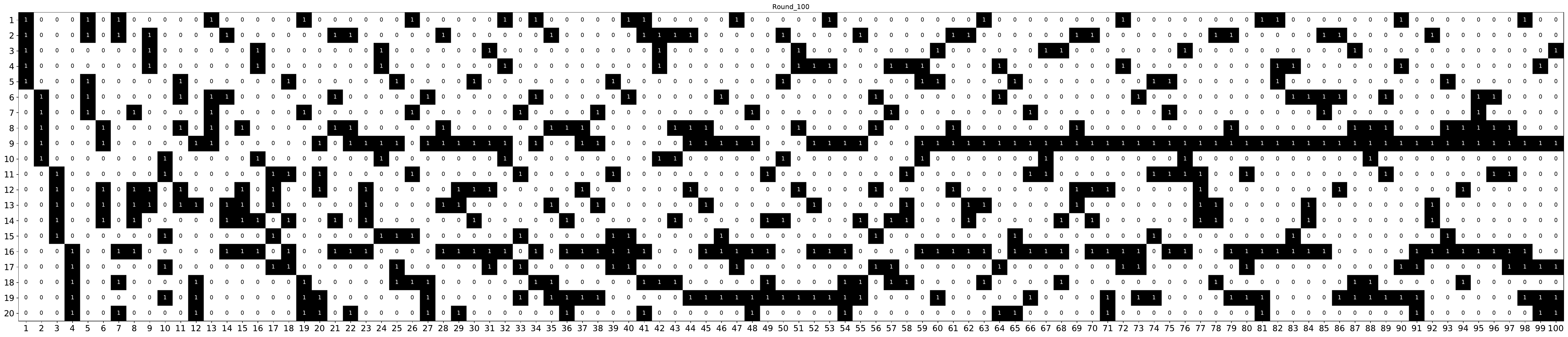}
        \caption{UCB-IID}
        \label{fig:ucb-IID}
    \end{subfigure}
    \caption{Client Selection in IID Case}
    \label{fig:iid}
\end{figure*}

We verified the effectiveness of client selection based on the ucb algorithm. As a comparison, we chose a random strategy method for comparison. Under different data distribution conditions, 20 clients were used to perform 100 global trainings. Five clients were selected in each training round to participate in the training. Figures \ref{fig:non-iid} and \ref{fig:iid} show the client participation of the two selection strategies under Non-iid and iid distributions. Regardless of the data distribution, the Random strategy selects each client more evenly, which is not conducive to the participation of clients with greater contributions in training. UCB reflects the trend of more work and more pay. At the same time, it does not over-converge on some clients, which will help the model not to over-converge on some clients and improve the generalization ability of the model.
\subsection{Contribution Method Assessment}
\begin{figure*}[!t]
 \centering
    \begin{subfigure}[!t]{0.3\textwidth}
        \includegraphics[width=1.0\columnwidth]{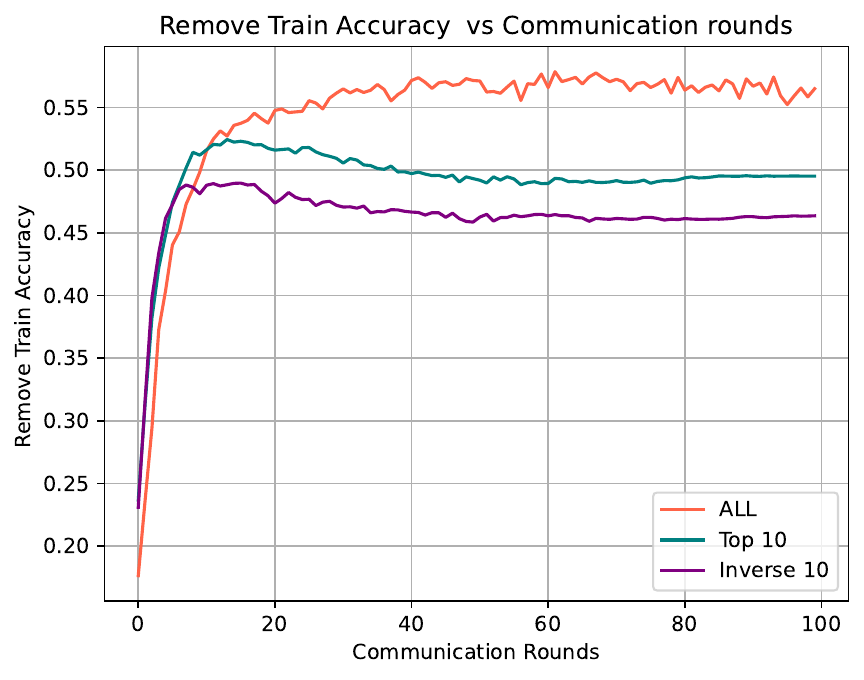}
        \caption{CIFAR}
        \label{fig:cifar}
    \end{subfigure}
    \begin{subfigure}[!t]{0.3\textwidth}
        \includegraphics[width=1.0\columnwidth]{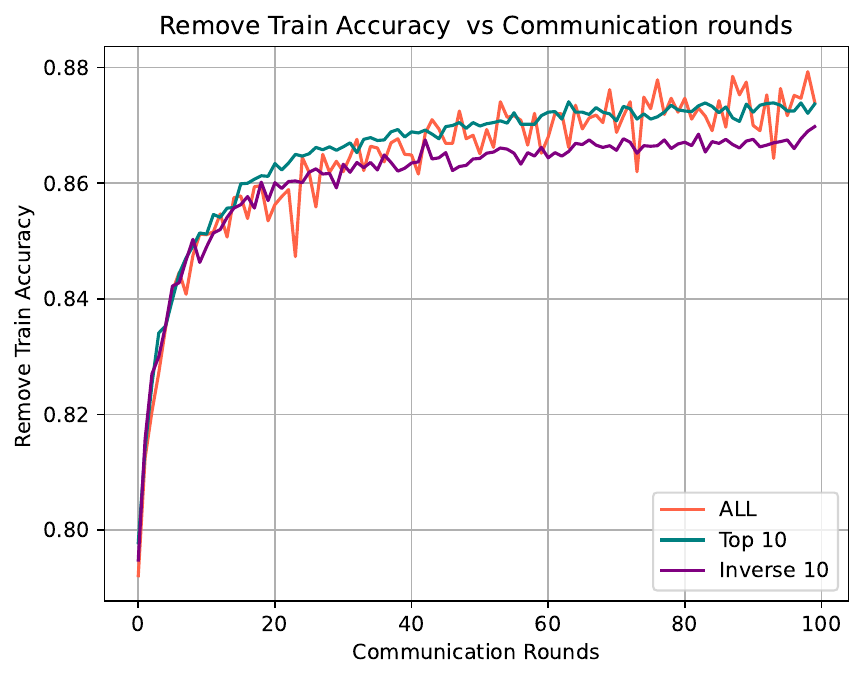}
        \caption{FMNIST}
        \label{fig:fmnist}
    \end{subfigure}
    \begin{subfigure}[!t]{0.3\textwidth}
        \includegraphics[width=1.0\columnwidth]{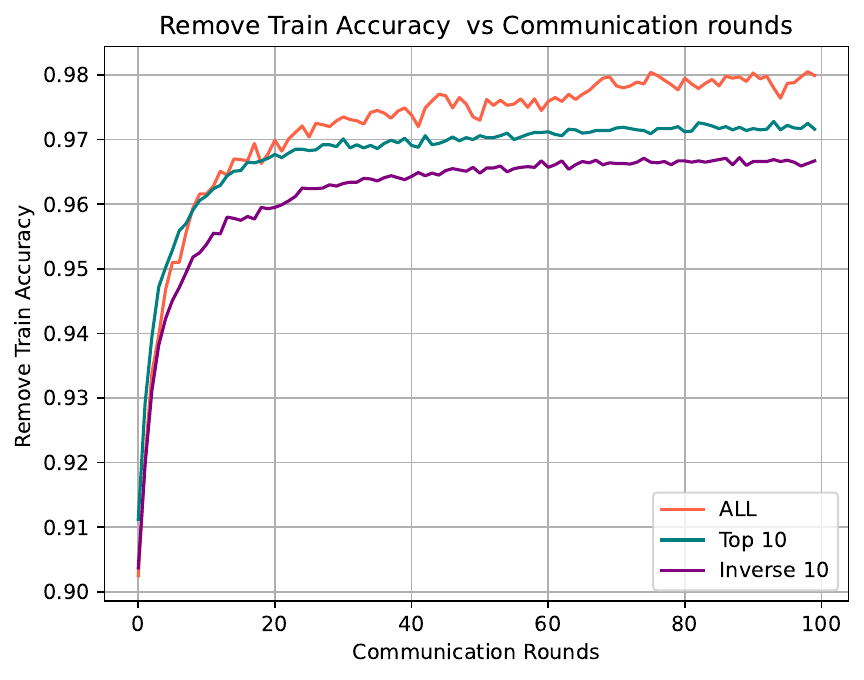}
        \caption{MNIST}
        \label{fig:mnist}
    \end{subfigure}
    \begin{subfigure}[!t]{0.3\textwidth}
        \includegraphics[width=1.0\columnwidth]{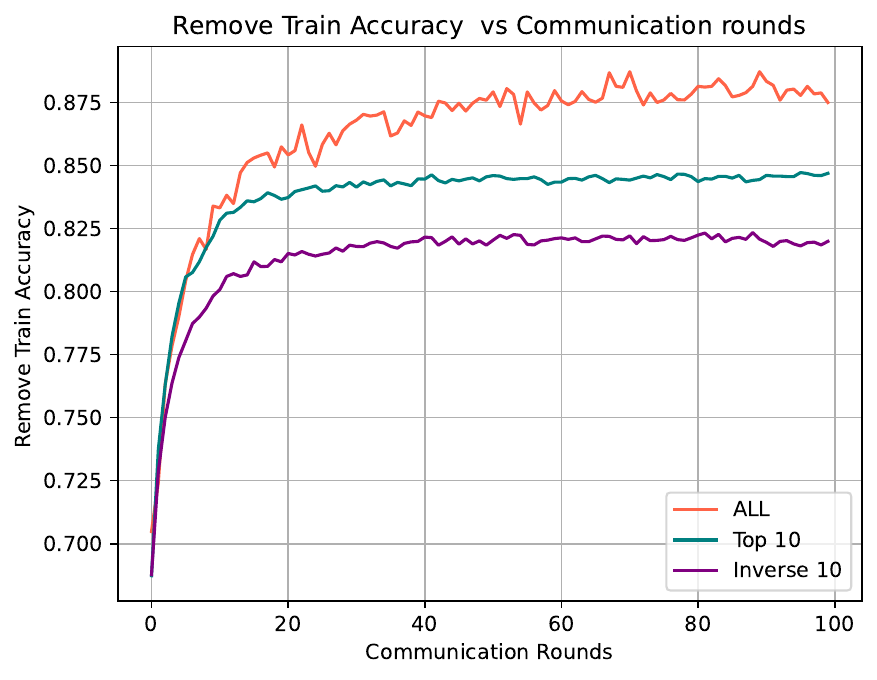}
        \caption{KMNIST}
        \label{fig:kmnist}
    \end{subfigure}
    \begin{subfigure}[!t]{0.3\textwidth}
        \includegraphics[width=1.0\columnwidth]{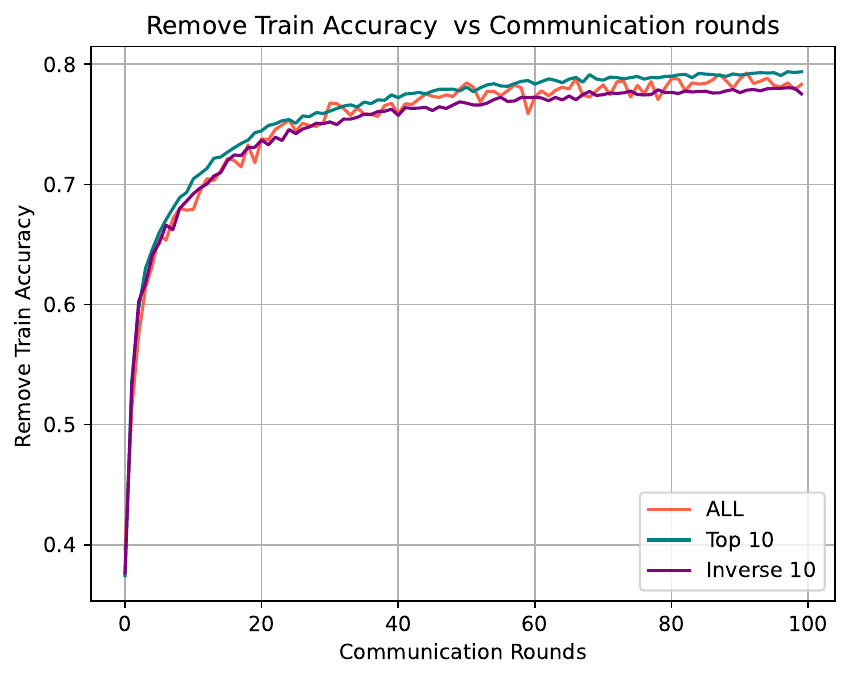}
        \caption{SVHN}
        \label{fig:svhn}
    \end{subfigure}
    \caption{Model training caused by removing some clients}
    \label{fig:remove}
\end{figure*}

 To verify the effectiveness of the contribution evaluation method proposed in this work, we selected 50 clients to conduct 100 rounds of global training first, and obtained the contribution value of each client in this process. Then, we sorted the 50 clients in descending order by contribution value, and selected the top 10 and bottom 10 clients respectively, that is, removed the clients ranked in the [10:50] and [1:40] intervals and re-trained them, and recorded the training accuracy of these three groups of clients. At the same time, to verify the universality of this method under different data sets, we selected 5 different data sets for experiments. As shown in Figure \ref{fig:remove}, there is a significant gap in accuracy between the top 10 clients and the bottom 10 clients. On the svhn and fmnist data sets, the training effect and smoothness of the top 10 clients are even better than when all clients participate. This shows that our method can effectively evaluate the contribution of clients, and the selected high-quality clients are representative.
\subsection{Client Participation Performance}
\begin{figure}[!t]
    \includegraphics[width=1.0\columnwidth]{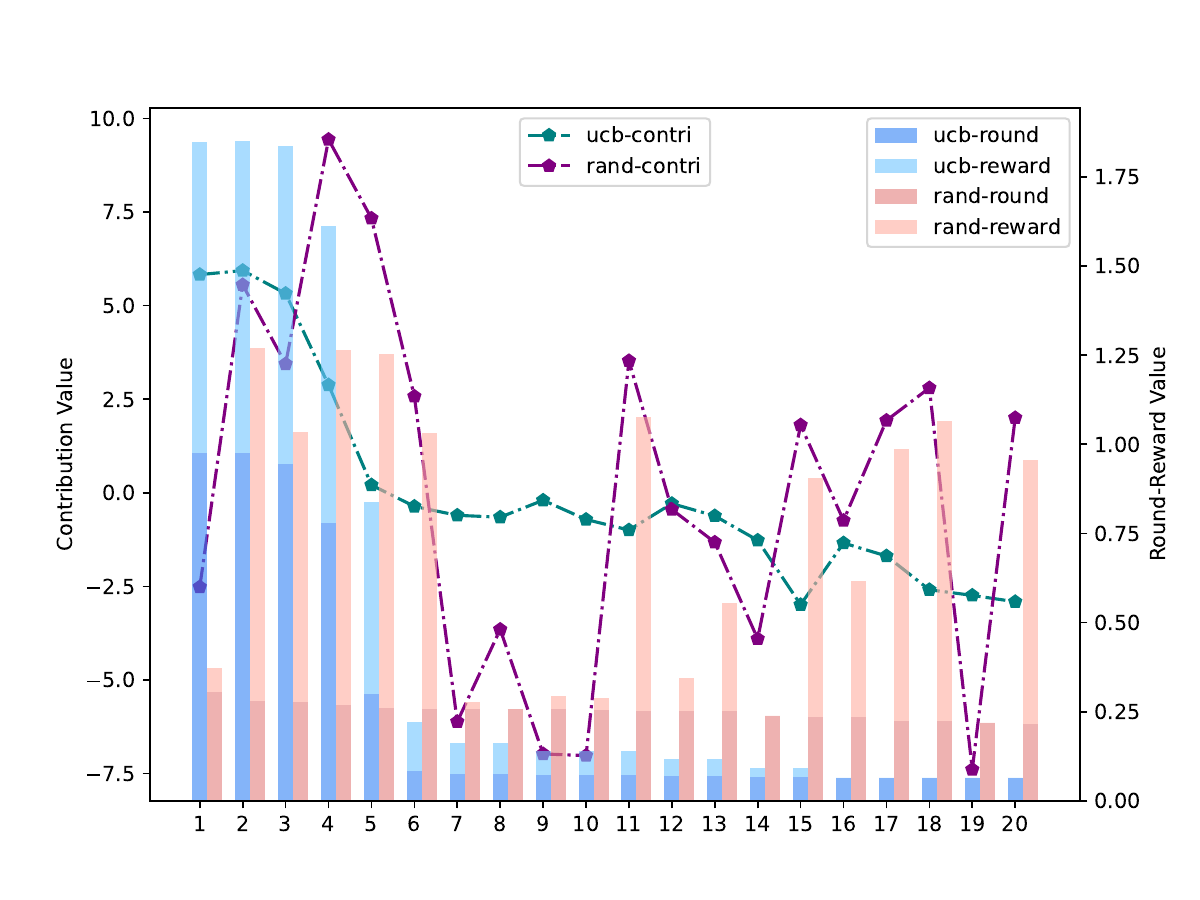}
    \caption{Client Participation Performance}
    \label{fig:comp-result}
\end{figure}

 Based on the ucb strategy, we compared the performance of 20 clients under the Random strategy after 100 rounds of global training, including the client's contribution value and reward probability-participation rounds. We normalized the participation rounds of each client, and the reward probability was obtained by dividing the reward value of each round in the reward function designed in this work by the total rounds, which represents the probability of selecting a client to make a positive contribution, as shown in Figure \ref{fig:comp-result}. The broken line in the figure represents the contribution value of the client, the light-colored part of the bar chart represents the reward probability of the client, and the dark-colored part represents the participation rounds of the client. Under the ucb strategy, the higher the reward probability, the more participation rounds the client gets, thereby obtaining a higher contribution value. When the contribution value of the client is negative, the client will participate in very few rounds, which shows the rationality of our reward function design, which can accurately reflect the client's performance, so that clients with positive contributions can get more participation opportunities. Under the Random strategy, we also record the reward probability of each client. Even though the reward probability correctly reflects the client's performance, due to the average nature of random selection, each client can get a similar number of participation rounds regardless of its performance. This means that clients with good performance do not get more opportunities to participate, while clients with poor performance may get more opportunities, which affects the accuracy of the model and cannot guarantee good contribution fairness.
	\subsection{Model Accuracy Assessment}
\begin{figure*}[!t]
    \centering
    \begin{subfigure}{0.3\linewidth}
        \includegraphics[width=\linewidth]{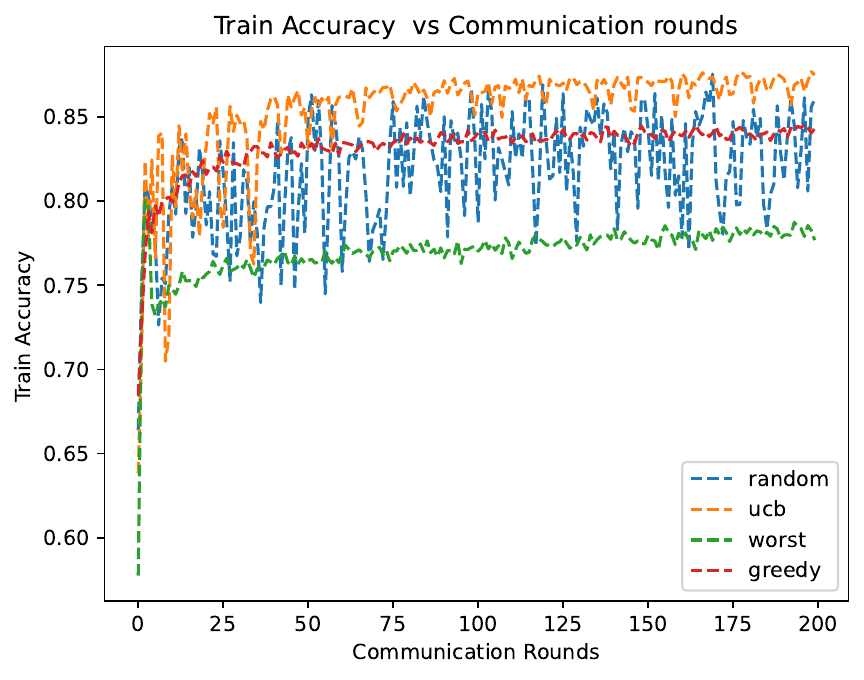}
        \caption{fmnist non-iid}
        \label{fig:image1}
    \end{subfigure}
    \begin{subfigure}{0.3\linewidth}
        \includegraphics[width=\linewidth]{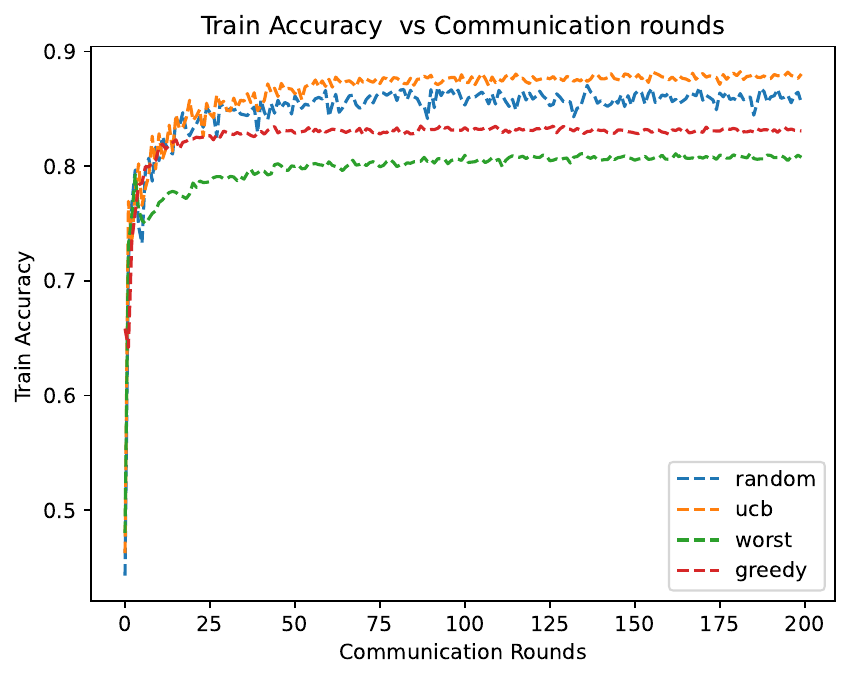}
        \caption{kmnist non-iid}
        \label{fig:image2}
    \end{subfigure}
    \begin{subfigure}{0.3\linewidth}
        \includegraphics[width=\linewidth]{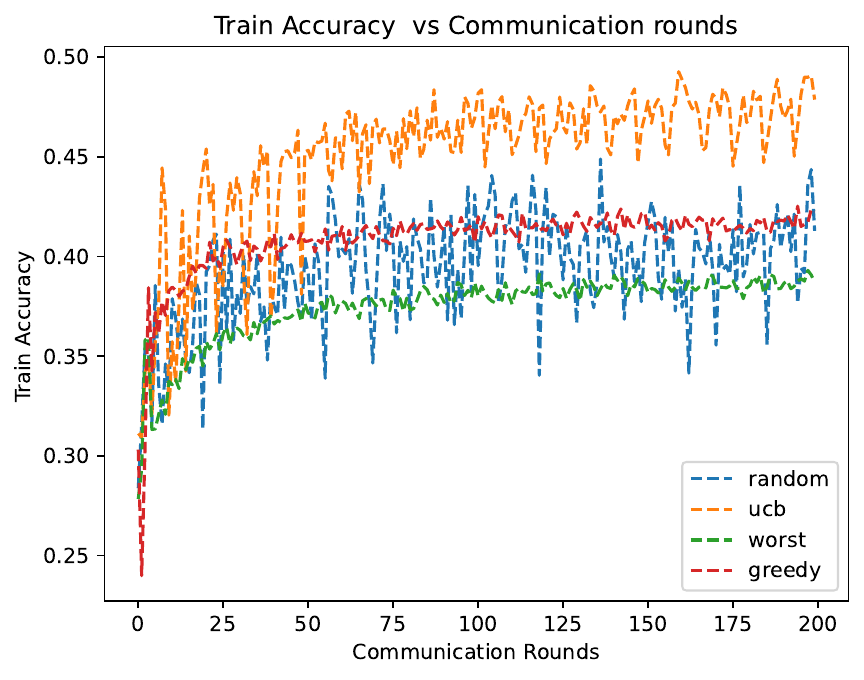}
        \caption{cifar-10 non-iid}
        \label{fig:image3}
    \end{subfigure}
    \begin{subfigure}{0.3\linewidth}
        \includegraphics[width=\linewidth]{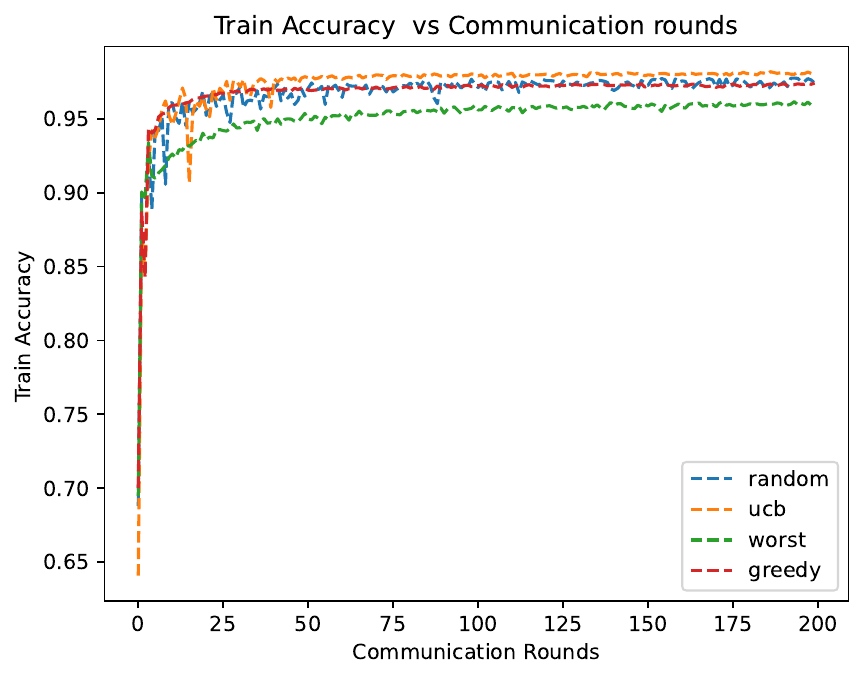}
        \caption{mnist non-iid}
        \label{fig:image4}
    \end{subfigure}
    \begin{subfigure}{0.3\linewidth}
        \includegraphics[width=\linewidth]{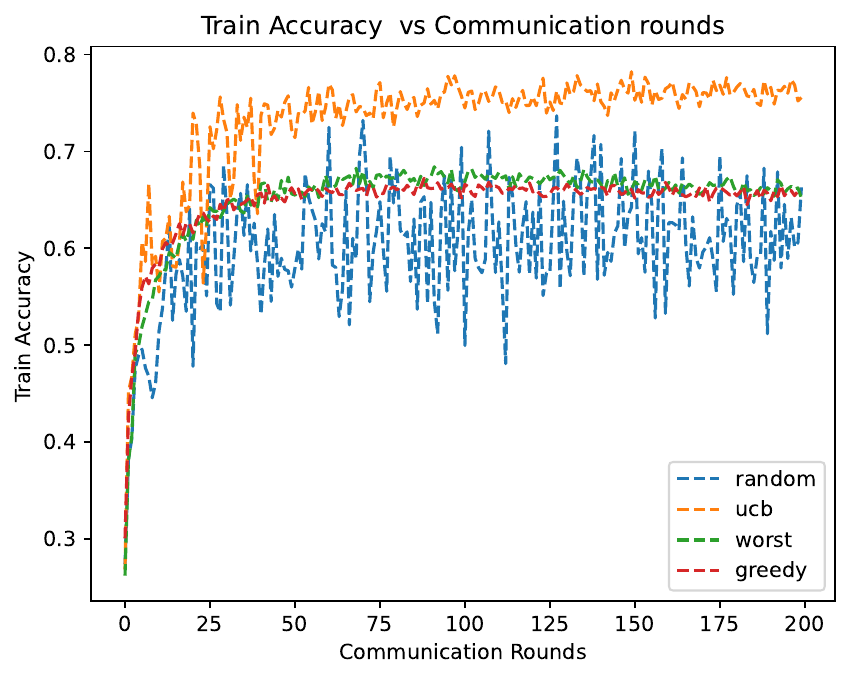}
        \caption{svhn non-iid}
        \label{fig:image5}
    \end{subfigure}
    \caption{Non-IID Training Accuracy}
    \label{fig:noniid-train-acc}
\end{figure*}

	To verify the effectiveness of strategy selection, we compared the training accuracy of several strategy selections under five different data sets, including the ucb-based strategy proposed in this work, the random selection strategy random, the worst selection strategy worst: that is, always select the client with the worst performance under the ucb-based strategy, and the greedy strategy greedy: greedily select the client with the most rounds of participation. In each round of training, we choose 5 out of 20 clients for training. At the same time, we list the test accuracy after the training.
	\begin{itemize}
		\item As shown in Table \ref{table:iid_test_acc} and Figure \ref{fig:noniid-train-acc}, the training accuracy under different data distributions, it can be observed that the training accuracy based on the UCB strategy is better than other strategies in different data distributions and different data sets. The gap is more obvious on the more complex the data set, and the training accuracy curve of the worst strategy is always at the bottom, which also shows the effectiveness of our strategy selection. The greedy strategy is equivalent to the worst strategy in most cases and converges to low accuracy too early. This shows that the participation round cannot reflect the performance of the client. Greedy selection of clients with more participation rounds will make the model converge to some clients, resulting in low model accuracy.
    \begin{table}[!t]
    \renewcommand\arraystretch{2}
    \centering
    \begin{tabular}{|c|c|c|c|c|c|}
        \hline & \textbf{FMNIST} & \textbf{KMNIST} & \textbf{Cifar-10} & \textbf{MNIST} & \textbf{SVHN}\\
        \hline
        UCB & 79.42 & 84.09 & 50.26 & 86.23 & 84.1195 \\
        \hline
        Random & 78.45 & 78.86 & 36.93 & 80.94 & 60.7675 \\
        \hline
        Greedy & 69.64 & 71.61 & 36.27 & 82.76 & 76.9169 \\
        \hline
        Worst & 44.72 & 58.26 & 26.26 & 31.76 & 44.4914 \\
        \hline
    \end{tabular}
    \caption{IID Test Accuracy}  
    \label{table:iid_test_acc}
\end{table}
\begin{figure*}[!t]
    \centering
    \begin{subfigure}{0.3\linewidth}
        \includegraphics[width=\linewidth]{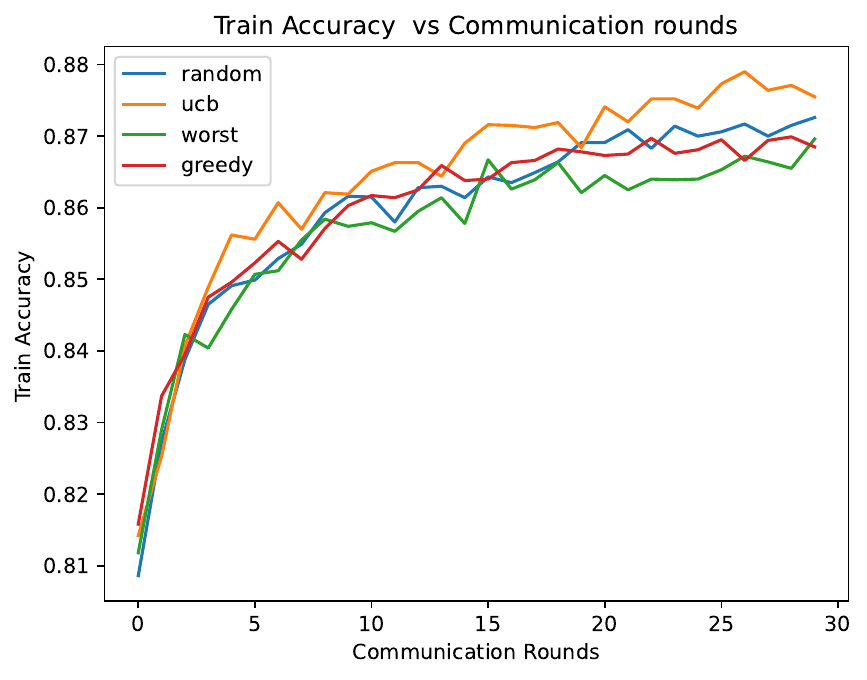}
        \caption{fmnist iid}
        \label{fig:image1}
    \end{subfigure}
    \begin{subfigure}{0.3\linewidth}
        \includegraphics[width=\linewidth]{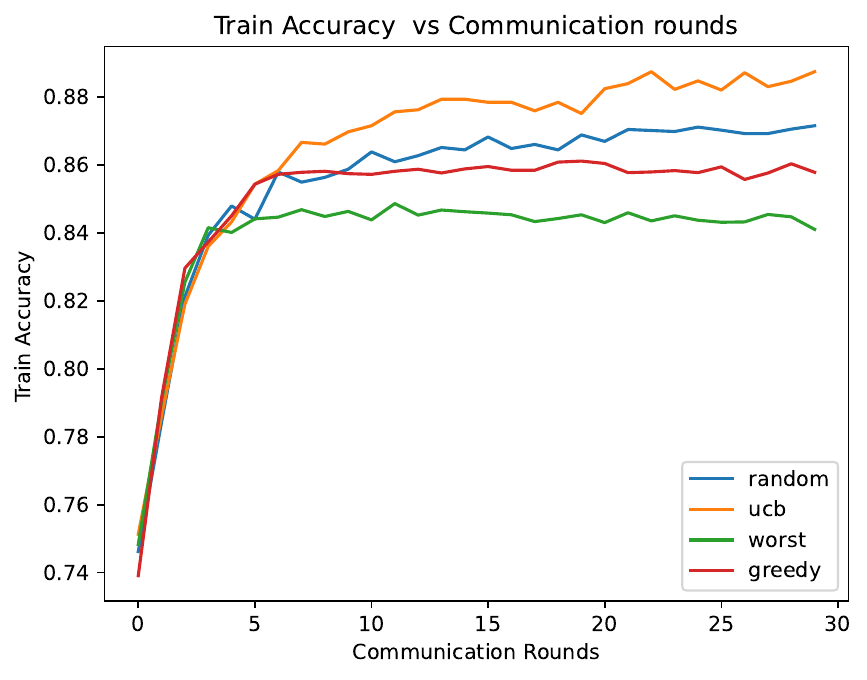}
        \caption{kmnist iid}
        \label{fig:image2}
    \end{subfigure}
    \begin{subfigure}{0.3\linewidth}
        \includegraphics[width=\linewidth]{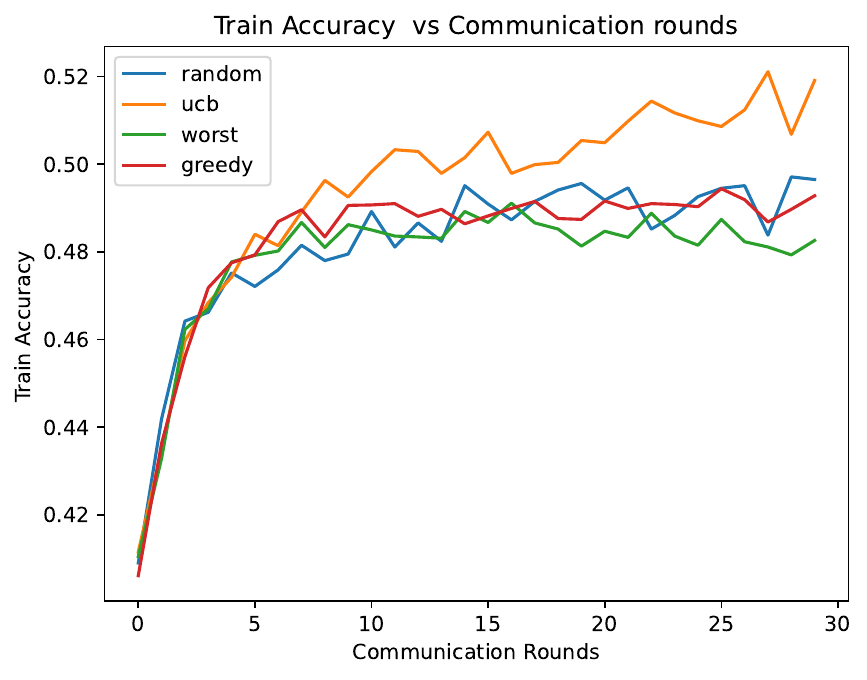}
        \caption{cifar-10 iid}
        \label{fig:image3}
    \end{subfigure}
    \begin{subfigure}{0.3\linewidth}
        \includegraphics[width=\linewidth]{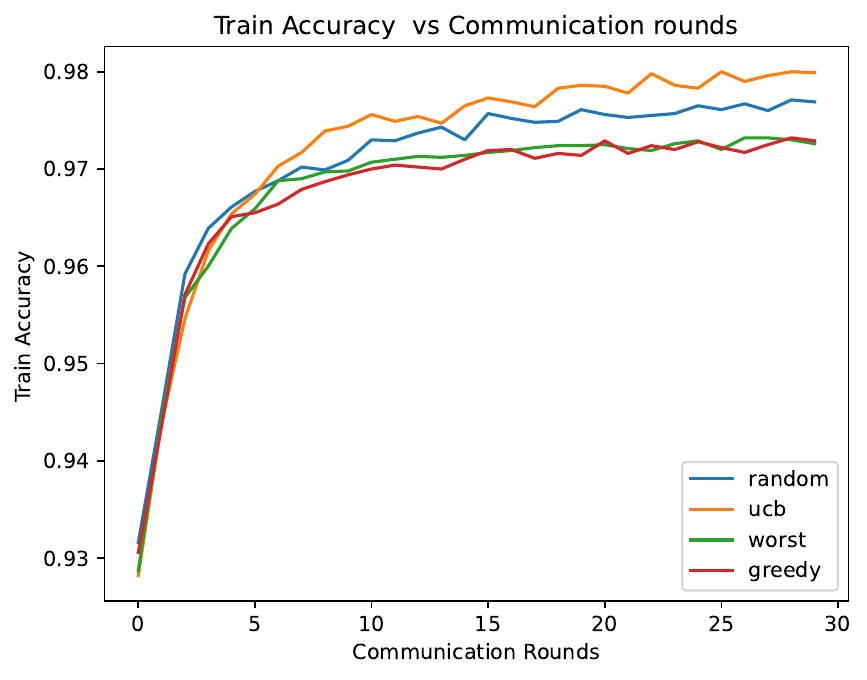}
        \caption{mnist iid}
        \label{fig:image4}
    \end{subfigure}
    \begin{subfigure}{0.3\linewidth}
        \includegraphics[width=\linewidth]{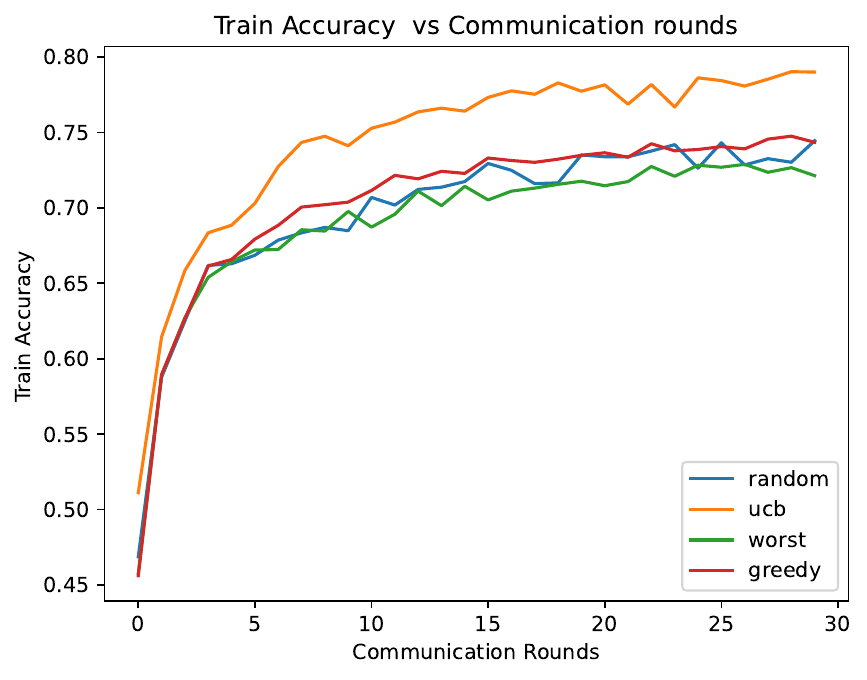}
        \caption{svhn iid}
        \label{fig:image5}
    \end{subfigure}
    \caption{IID Training Accuracy}
    \label{fig:iid-train-acc}
\end{figure*}
        \item As shown in Tables \ref{table:noniid_test_acc} and Figure \ref{fig:iid-train-acc}, the test accuracy of the model trained based on the UCB strategy is better than other selection strategies on each different data set. On more complex data sets, such as CIFARR-10 and SVHN, the accuracy gap is more obvious, even exceeding 20 percentage points, which shows that our strategy selection method can perform better when dealing with more complex tasks.
 \end{itemize}
\begin{table}[!t]
    \renewcommand\arraystretch{2}
    \centering
    \begin{tabular}{|c|c|c|c|c|c|}
        \hline & \textbf{FMNIST} & \textbf{KMNIST} & \textbf{Cifar-10} & \textbf{MNIST} & \textbf{SVHN}\\
				\hline   UCB & 87.48 & 88.05 & 47.86 & 98.17 & 75.5762 \\
				\hline   Random & 86.01 & 85.53 & 41.28 & 96.42 & 66.2953 \\
				\hline   Greedy & 84.30 & 83.08 & 38.71 & 96.43 & 65.8920 \\
				\hline   Worst & 77.67 & 80.78 & 42.24 & 95.07 & 65.1890 \\
				\hline
			\end{tabular}   
    \caption{Non-IID Test Accuracy} 
    \label{table:noniid_test_acc}
\end{table}
	\subsection{Evaluate contribution aggregation function}
\begin{figure*}[!t]
     \centering
    \begin{subfigure}[!t]{0.23\linewidth}
        \includegraphics[width=\linewidth]{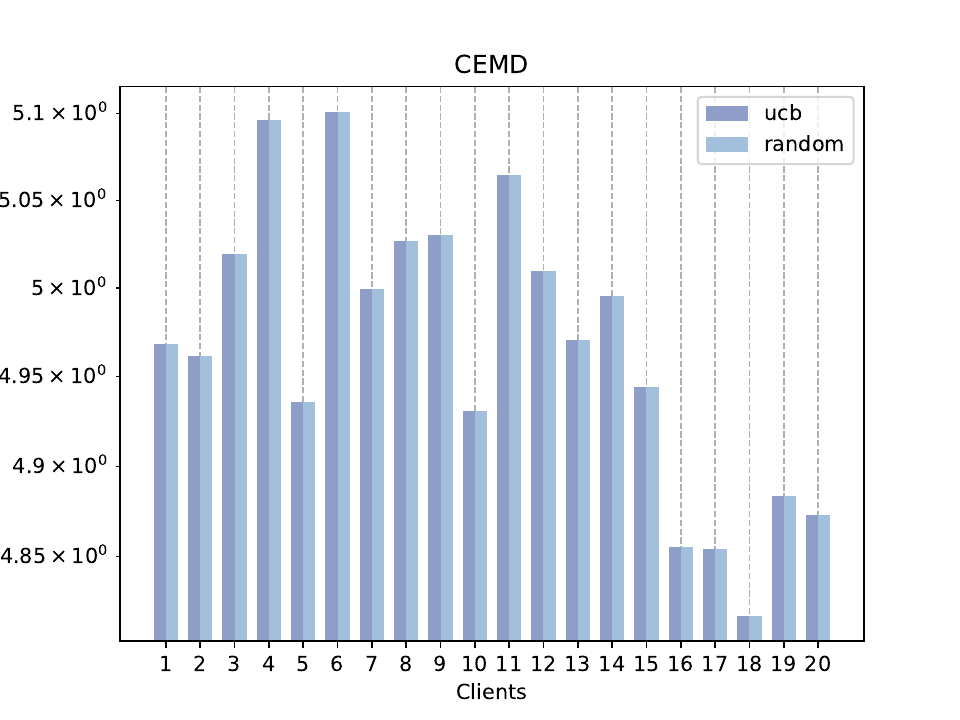}
        \caption{}
        \label{fig:CEMD}
    \end{subfigure}
    \begin{subfigure}[!t]{0.23\linewidth}
        \includegraphics[width=\linewidth]{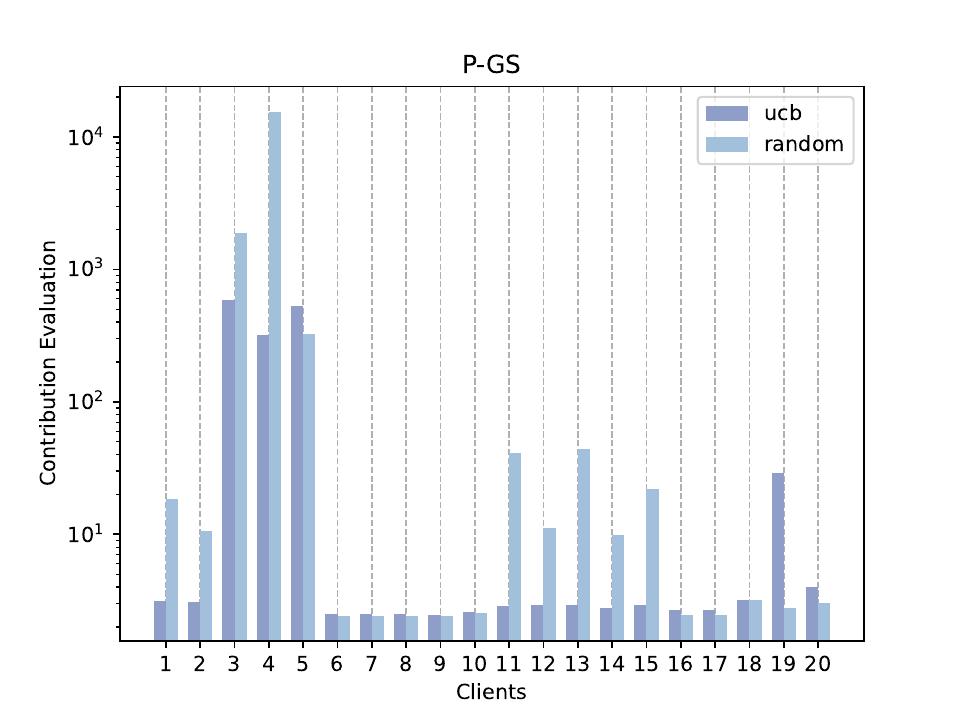}
        \caption{}
        \label{fig:P-GS}
    \end{subfigure}
    \begin{subfigure}[!t]{0.23\linewidth}
        \includegraphics[width=\linewidth]{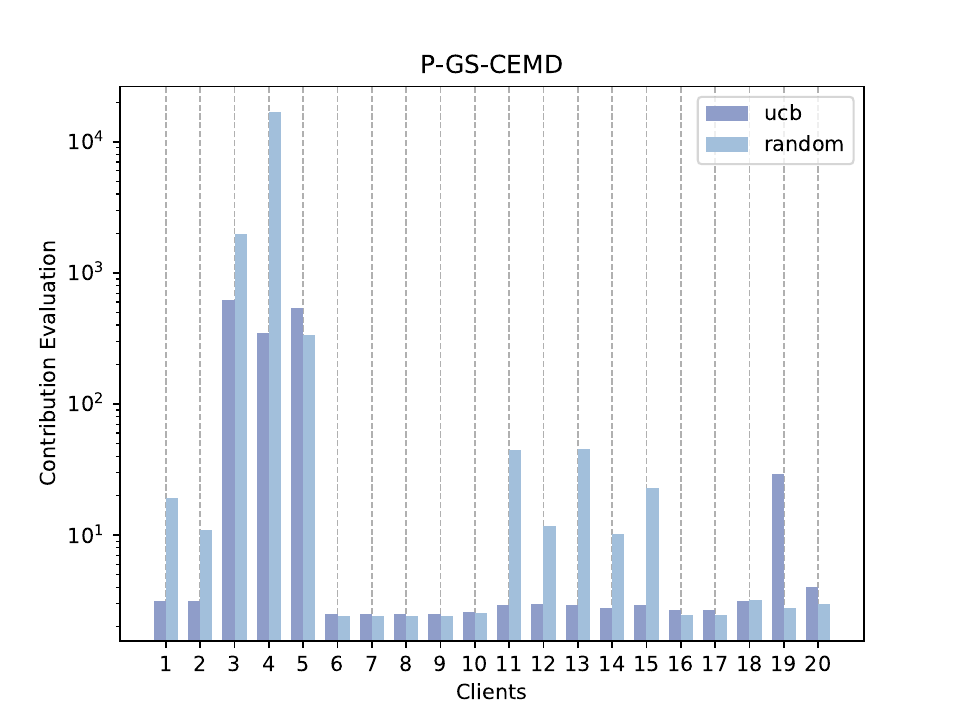}
        \caption{}
        \label{fig:P-GS-CEMD}
    \end{subfigure}
    \begin{subfigure}[!t]{0.23\linewidth}
        \includegraphics[width=\linewidth]{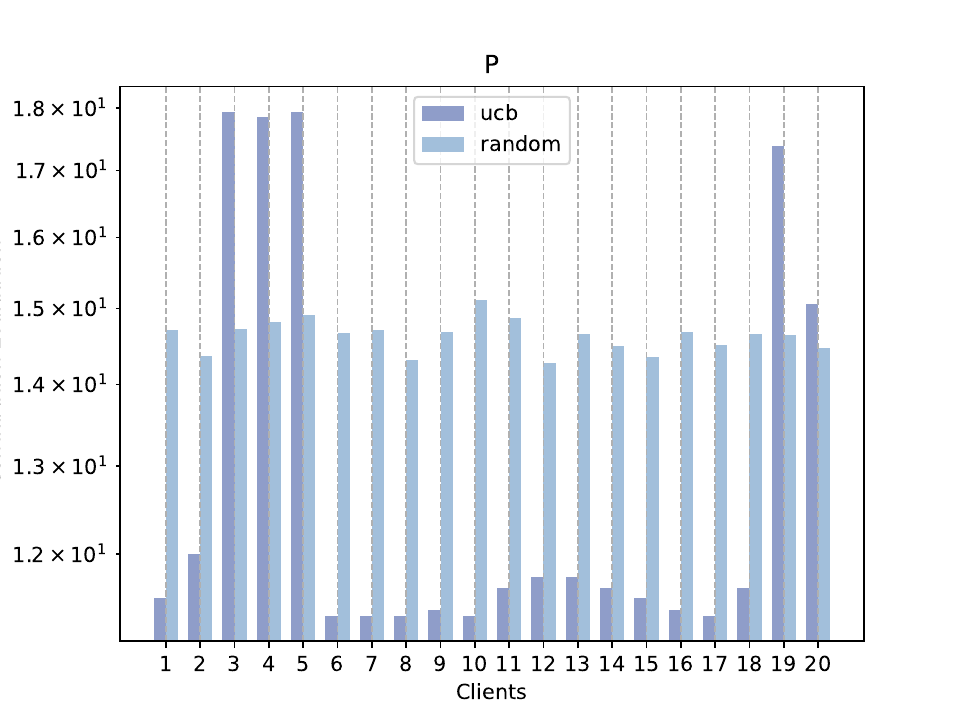}
        \caption{}
        \label{fig:image4}
    \end{subfigure}
    \begin{subfigure}[!t]{0.23\linewidth}
        \includegraphics[width=\linewidth]{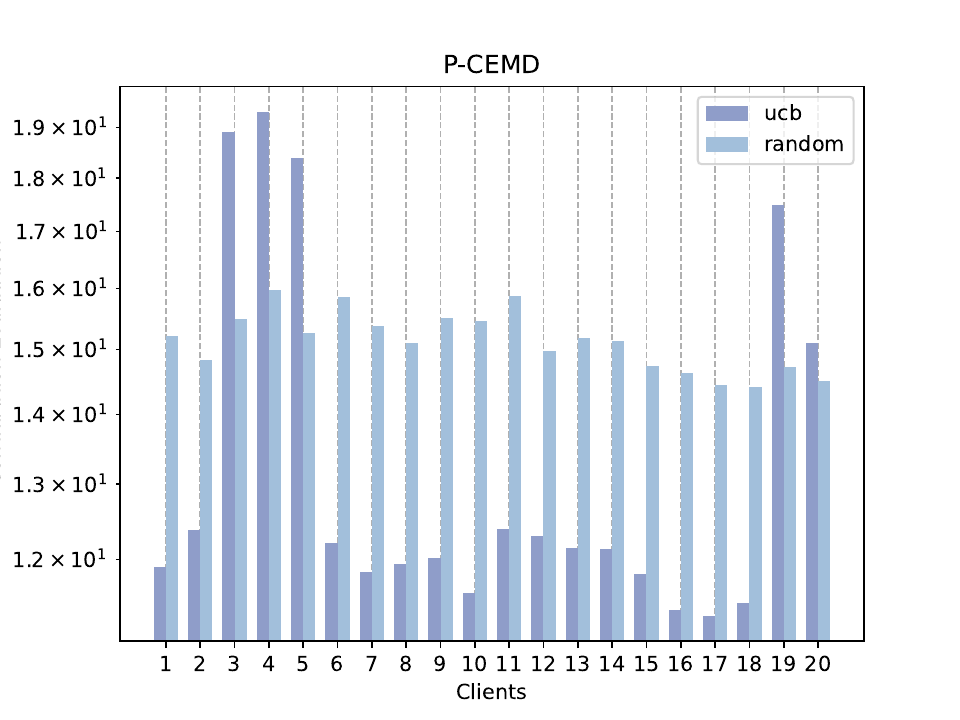}
        \caption{}
        \label{fig:P-CEMD}
    \end{subfigure}
    \begin{subfigure}[!t]{0.25\linewidth}
        \includegraphics[width=\linewidth]{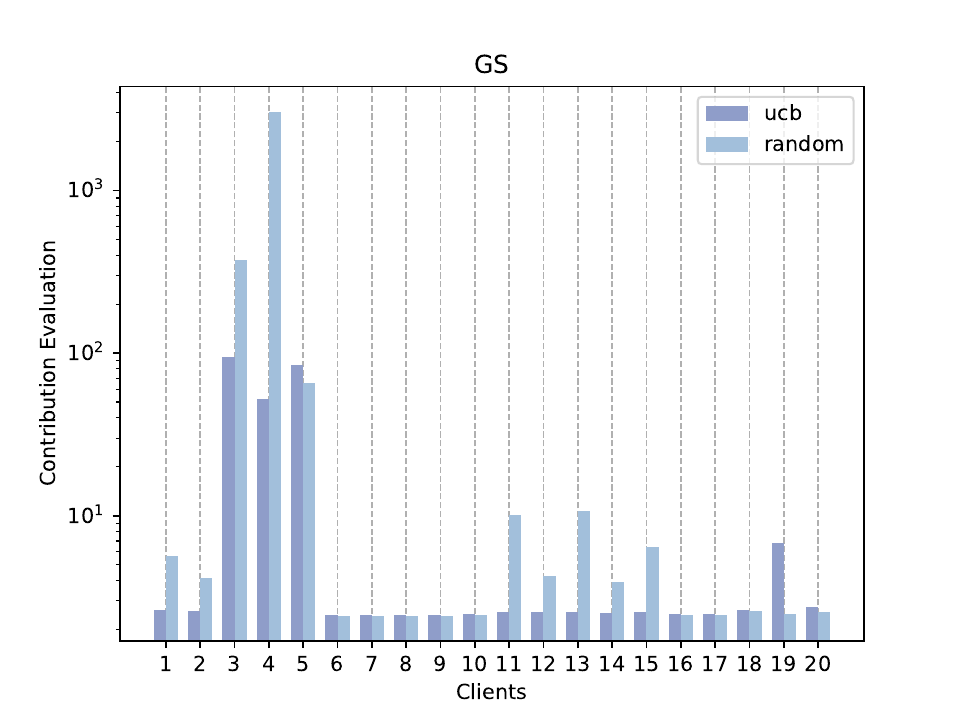}
        \caption{}
        \label{fig:GS}
    \end{subfigure}
    \begin{subfigure}[!t]{0.25\linewidth}
        \includegraphics[width=\linewidth]{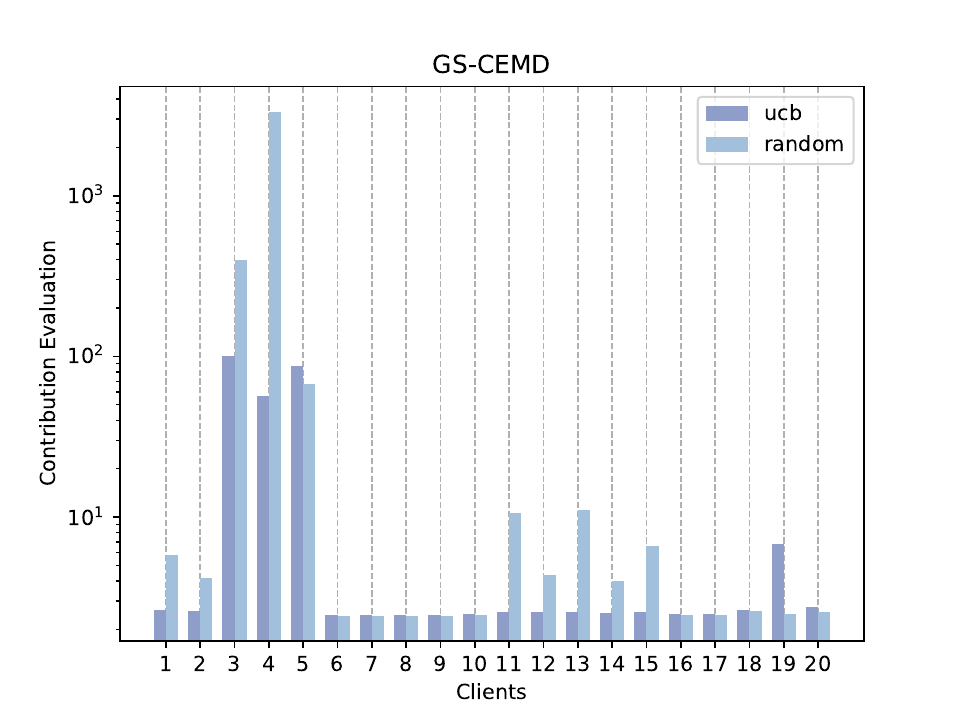}
        \caption{}
        \label{fig:GS-CEMD}
    \end{subfigure}
    \caption{Contribution aggregation function that retains specific parameters}
    \label{fig:Parameter Function}
\end{figure*}

\begin{figure}[!t]
    \includegraphics[width=1.0\linewidth]{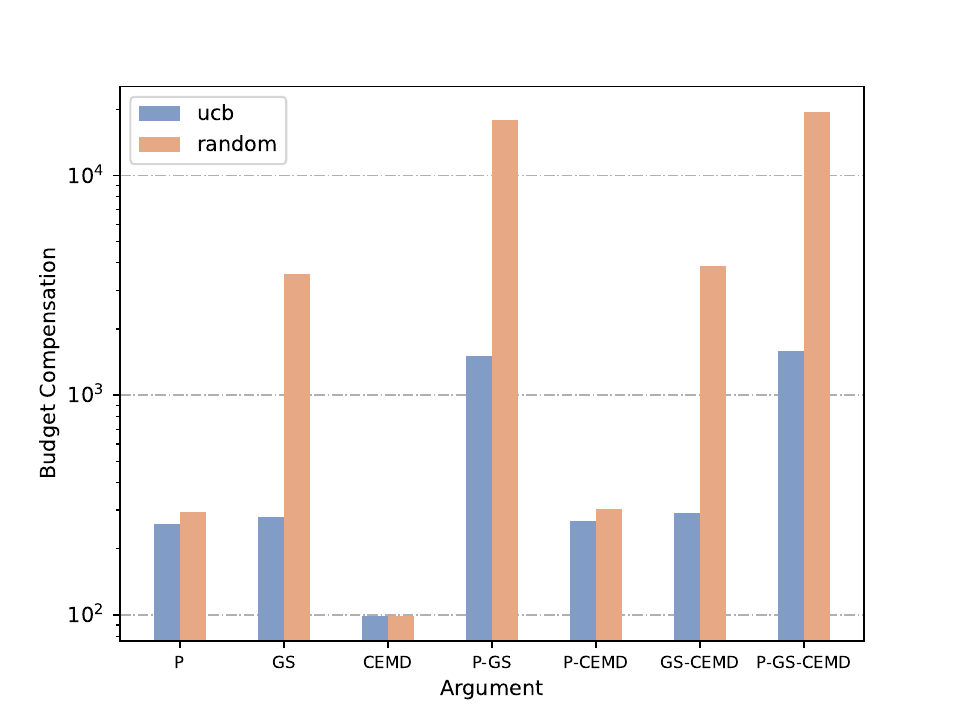}
    \caption{Budget compensation under specific parameter combination}
    \label{fig:budget-comp-result}
\end{figure}

	To verify the rationality of the contribution aggregation function and compensation design, we added the parameters of the function one by one and observed the impact of the retained parameters on the function value. As shown in Figure \ref{fig:Parameter Function}, when the CEMD parameter is retained, the data distribution between different clients is obviously reflected. When only this parameter is retained, there is no distinction under different strategies, which is not suitable for separate measurement. When the P parameter is retained, it will be affected by the selection strategy. For example, the average of the training rounds under the Random strategy leads to almost no difference in the function value dominated by P. Only when GS is retained or other parameters are combined with GS can the specific contribution of the client be better reflected, and the function design makes the function value approximately proportional to the GS value. When the three parameters are combined, the function value will be improved as a whole. At the same time, we calculated the budget compensation under different parameter combinations. As shown in the Figure \ref{fig:budget-comp-result}. Under the ucb strategy, the budget compensation expenditures obtained by different parameters are lower than those of the Random strategy, thus achieving the purpose of low cost for agents. The compensation design is the normalized value of contribution aggregation, so P-GS-CEMD can reflect that clients with greater contributions can achieve high income.
	\section{Conclusion}\label{sec:conclusion}
In this paper, we analyze the needs between sellers and agents in the data trading market, propose a data trading framework based on the federated learning architecture, design a seller selection algorithm and a contribution evaluation method as well as a compensation method. Through experimental simulation results, we prove the rationality and effectiveness of the framework. At the same time, we achieve the goal of low cost for agents and high income for sellers, which helps to solve the problems existing in the data trading market and promote the development of this field. In future work, considering the privacy and timeliness of data is an aspect worth studying.

\end{document}